\documentclass{article}
\usepackage{PRIMEarxiv}
\usepackage[T1]{fontenc}
\usepackage{graphicx}
\usepackage{hyperref}    

\usepackage[utf8]{inputenc}
\usepackage{amsmath}
\usepackage{bbm}
\usepackage{cleveref}
\usepackage{comment}
\usepackage{url}
\usepackage{mathtools}
\usepackage{fancyhdr}
\usepackage{framed}
\usepackage{amsfonts}
\usepackage{amssymb}
\usepackage{enumerate}
\usepackage{color}
\usepackage{caption}
\usepackage{subcaption}
\usepackage[misc]{ifsym}
\usepackage{enumitem}
\usepackage{booktabs}
\usepackage{multirow}
\usepackage{cite}  

\makeatletter
\newcommand{\pfnsymbol}[1]{%
  \textsuperscript{\@fnsymbol{#1}}%
}
\makeatother

\title{What's Wrong with Your Synthetic Tabular Data? Using Explainable AI to Evaluate Generative Models}

\author{
Jan Kapar$^{1,2}$, Niklas Koenen$^{1,2}$, Martin Jullum$^{3}$
\and
$^1$Leibniz Institute for Prevention Research \& Epidemiology – BIPS, Germany
\and
$^2$Faculty of Mathematics and Computer Science, University of Bremen, Germany 
\and
$^3$Norwegian Computing Center, Oslo, Norway\\
\texttt{jullum@nr.no} \\
}

\begin{document}

\maketitle

\begin{abstract}
Evaluating synthetic tabular data is challenging, since they can differ from the real data in so many ways. There exist numerous metrics of synthetic data quality, ranging from statistical distances to predictive performance, often providing conflicting results. Moreover, they fail to explain or pinpoint the specific weaknesses in the synthetic data. To address this, we apply explainable AI (XAI) techniques to a binary detection classifier trained to distinguish real from synthetic data. While the classifier identifies distributional differences, XAI concepts such as feature importance and feature effects, analyzed through methods like permutation feature importance, partial dependence plots, Shapley values and counterfactual explanations, reveal \emph{why} synthetic data are distinguishable, highlighting inconsistencies, unrealistic dependencies, or missing patterns. This interpretability increases transparency in synthetic data evaluation and provides deeper insights beyond conventional metrics, helping diagnose and improve synthetic data quality. We apply our approach to two tabular datasets and generative models, showing that it uncovers issues overlooked by standard evaluation techniques.

\end{abstract}

\section{Introduction}
\label{sec:intro}
The rapid development of generative modeling, also known as generative artificial intelligence (GenAI), is driving profound changes in business, science, education, creative processes, and our everyday lives. While the maturation of transformer-based architectures \cite{vaswani2017attention} and diffusion models \cite{ho2020denoising,song2020score} has led to previously unimaginable possibilities and quality leaps in text, image, audio, and video generation, the advances in tabular data synthesis lag behind. However, generative modeling has great potential for tabular data: A large amount of data in organizations, research and medicine is organized in a tabular format, and there is a wide range of useful applications, such as privacy-preserving data sharing, data augmentation and balancing, missing data imputation and what-if analyses \cite{jordon2022synthetic}.

What is it that makes tabular data synthesis so difficult for methods which excel in image and text domains? How can we examine which exact parts of a synthetic dataset contain implausible values or patterns? Answering these questions is not trivial, as already measuring the quality of synthetic data is not straightforward: As opposed to supervised learning, there are no direct performance measures for generative modeling as a mainly unsupervised discipline through the absence of labels. 
Moreover, even in relatively low dimensions, dependencies between features -- of numerical or categorical nature -- are hard to comprehend for humans. This makes the evaluation of tabular data synthesis quality even more challenging compared to image or text data where human experts can assess the results more easily. 
There are many different concepts for evaluating the performance of generative models, and not all of them are available for every model and data type. New measures are also frequently proposed to address the shortcomings of the previous ones. 
However, transparency regarding the underlying \textit{reasons} for poorer synthesis quality, reflected in lower performance scores, is rarely available.
Furthermore, while some of the measures allow for evaluation at individual observation level \cite{lopez2022revisiting,alaa2022faithful}, it often remains unclear which specific feature values or combinations are responsible for the poor quality of a specific synthetic observation. (See \Cref{subsec:background_genmodel} for more background about generative models and their evaluation.)

Explainable artificial intelligence (XAI), also referred to as interpretable machine learning (IML), tries to explain the outputs and decision-making of machine learning models. This is an emerging machine learning discipline, as high performing machine learning models are often black boxes due to their complexity, and there is an urgent demand for methods to make them more transparent and easier to trust. Some XAI methods are able to attribute to each feature - or even feature interaction - its contribution to a model's output. However, the majority of these XAI methods is designed for supervised learning and not directly applicable to generative modeling. (See \Cref{subsec:background_IML} for more background about XAI.)

\paragraph{Contributions} We propose to leverage a synthetic data detection model to evaluate the performance of a generative model and to use it as a supervised proxy model in order to obtain more detailed insights about the strengths and weaknesses of the synthetic data via common XAI methods. While the detection model itself can give an indication of the overall and individual discernibility of real and synthetic data and about the fidelity and diversity of the synthetic data (\Cref{subsec:methods_detection}), we go one step further: As our main contribution, we provide a set of suitable global and local XAI tools which can be used to unlock more detailed insights by answering the following questions (\Cref{subsec:methods_IML}):

\begin{enumerate}[label=Q\arabic*.,leftmargin=*]
    \item \label{enum:Q1} Which features and feature dependencies were most challenging for the generative model?
    \item \label{enum:Q2} How do the generative models behave in low and high density areas of feature distributions? Which areas are under- or overrepresented in the synthetic data?
    \item \label{enum:Q3} Which features and feature dependencies/interactions contributed most to the detection of an individual real or synthetic observation?
    \item \label{enum:Q4} Which minimal changes to a correctly classified synthetic observation could be performed to make it look realistic?
\end{enumerate}

\noindent
We demonstrate the utility of this approach by answering these questions for real data examples (\Cref{sec:examples}). This work is primarily aimed at practitioners and researchers in the field of generative modeling, who want to gain deeper insights about the quality of their synthetic data or about the strengths and limitations of their generative models. Secondarily, within the XAI community, we want to raise awareness of the specific challenges in explainability for generative models that have not yet been adequately addressed.

\section{Related Work}
\label{sec:related-work}

The intersection of XAI and generative modeling has not yet been sufficiently researched. Schneider \cite{schneider2024explainable} underscores the demand for interpretability methods for generative models and gives a road map for this research direction but does not refer to explaining synthetic data quality explicitly. Several works have focused on finding meaningful representations of the latent space of generative models \cite{higgins2017beta,chen2016infogan,dombrowski2024trade} to better understand and control the data generation process or have specifically examined the explainability of the attention mechanism and transformers \cite{abnar2020quantifying,vig2019multiscale}. XAI has successfully been incorporated in the training process of generative models to increase their performance \cite{nagisetty2020xai,zhou2018activation}, but these approaches are specifically aimed at image data generation.

In the opposite direction, generative modeling has supported XAI methods to explain the inner workings of neural networks \cite{nguyen2016synthesizing}, create more human-understandable interpretations of model decisions \cite{dravid2022medxgan}, calculate conditional feature importances \cite{blesch2025conditional,agarwal2020explaining}, and produce more plausible counterfactuals \cite{nemirovsky2022countergan,dandl2024countarfactuals,redelmeier2024mcce}.

The following works are most closely related to our approach, though they emphasize different aspects and do not extend as far in our primary focus area of explaining synthetic data quality: Lopez-Paz et al. \cite{lopez2022revisiting} suggest to train a detection model to evaluate if two samples derive from the same distribution and analyze the statistical properties of such classifier two-sample tests. Zein et al. \cite{zein2022foolXGB} follow this strategy to demonstrate that machine learning utility as an evaluation measure may not be a reliable indicator for synthetic data quality. Both underline the interpretability of such classifier models and even perform a limited analysis using XAI, but do not focus on this aspect: Lopez-Paz et al. \cite{lopez2022revisiting} briefly mention several feature importance methods and show an example on image data, Zein et al. \cite{zein2022foolXGB} use impurity-based and permutation-based feature importances for synthetic tabular data detection. Neither of these works go beyond feature importance techniques which represent only an initial analysis step in the set of XAI tools presented in this work.

On image data, a wide range of existing studies use binary classifiers and their image-specific interpretation techniques for deep fake detection \cite{bird2024cifake,abir2023deepfakeImageXAI,baraheem2023AIvsAI}.

\section{Background}
\label{sec:background}

Since this work lies at the intersection of generative modeling and XAI and is intended for researchers and practitioners who do not necessarily have expertise in both disciplines, we provide a high-level overview of both fields.

\subsection{Generative Modeling of Tabular Data}
\label{subsec:background_genmodel}

Generative modeling or generative artificial intelligence is a machine learning sub-discipline concerned with generating realistic synthetic data: Let $D_\text{real}$ denote a dataset with instances from a feature space $\mathcal{X}$, each of which is a realization of a random variable $\mathbf{X}$. Given $D_\text{real}$, generative modeling tries to generate synthetic data $D_\text{syn}$ which follow the same joint distribution $p_{\mathbf{X}}$. As this original data distribution is unknown in real-world settings, generative modeling \textit{explicitly} or \textit{implicitly} approximates $p_{\mathbf{X}}$ in order to generate realistic instances and is therefore related to the discipline of joint density estimation \cite{jordon2022synthetic,mohamed2016implicit}:
Explicit generative models $G:\mathcal{X} \rightarrow \mathbb{R}_0^+$ directly model the underlying joint distribution ($G = \hat{p}_{\mathbf{X}}$) and provide a sampling routine to generate new instances. On the other hand, implicit generative models $G: \mathcal{L} \rightarrow \mathcal{X}$ are trained to generate realistic samples given random noise $\mathbf{z}$ from a latent-space $\mathcal{L}$ without being able to compute density likelihoods ($G(\mathbf{z}) \sim \hat{p}_{\mathbf{X}}$).

Early deep learning approaches such as variational autoencoders (VAEs) \cite{kingma2014auto}, generative adversarial networks (GANs) \cite{goodfellow2014generative}, and normalizing flows (NFs) \cite{rezende2015variational} have set the ground for the success of generative AI within the past decade. Modern transformer-based architectures \cite{vaswani2017attention}, denoising diffusion probabilistic models (DDPMs) \cite{ho2020denoising,song2020score}, often combined with autoregressive models (ARs) \cite{bengio2000neural}, are responsible for the recent hype about generative AI for realistic text, image, audio, and video synthesis \cite{bond2021deep}. 

While numerous adaptions of these methods for tabular data synthesis have been proposed \cite{xu2019ctgan,kotelnikov2023tabddpm,zhao2023tabula}, these often struggle to achieve the same overwhelming results as on image or text data. Tree-based generative methods \cite{nowok2016synthpop,watson2023adversarial} have demonstrated competitive or superior performance while being significantly less computationally demanding than deep learning approaches \cite{qian2023synthcity,fossing2024evaluationSynthcityModels}, which aligns to well-known findings from discriminative modeling \cite{grinsztajn2022tree,borisov2022deep,shwartz2022tabular}: This type of models showed the capacity to effectively deal with the challenges specific to tabular data, such as different feature types (e.g., continuous, discrete, categorical), odd-shaped feature distributions (e.g., multimodality, skew, truncation), complex feature dependencies, and lack of a natural positional or syntactical feature order as for image or text data. However, recent advances in deep learning for both discriminative and generative modeling seem to be promising steps towards closing the gap to tree-based methods \cite{hollmann2022tabpfn,zhangTabSyn}.

Another inherent challenge in tabular data synthesis is the assessment of its quality. Generative models in general lack direct evaluation methods, since they are usually trained in an unsupervised manner. An evaluation via test likelihoods as in density estimation is only available for explicit generative models and does not necessarily allow for implications about the synthetic data quality \cite{theis2016note}. On top of that, tabular data quality cannot innately be evaluated like data modalities such as images and text, which humans are able to process and judge naturally. Various evaluation concepts exist which try to cover different aspects of the complexity of tabular data: \emph{Fidelity} or \emph{precision} metrics measure the resemblance of a synthetic dataset to the original data, and \emph{coverage} or \emph{recall} its diversity \cite{sajjadi2018assessing,alaa2022faithful}. This is often measured by assessing how much of the synthetic data is covered by the original data distribution and vice versa. As a trivial synthesizer returning an exact copy of the original dataset would receive perfect scores in both previous metrics, a third dimension is often added: \emph{Authenticity}, \emph{overfitting} or -- closely related -- \emph{privacy} metrics measure the generalization or privacy preservation ability of the generative model \cite{alaa2022faithful,jordon2022synthetic}. The so-called \emph{detection score} or \emph{classifier two-sample test} gives a measure of how well binary machine learning classifiers can distinguish between original and synthetic data \cite{lopez2022revisiting,qian2023synthcity}. Alternatively, the distance between original and synthetic data distributions can be assessed using statistical tests and distance measures such as \emph{Kullback-Leibler divergence}, \emph{Wasserstein distance}, and \emph{maximum mean discrepancy} \cite{bischoff2024practical,mohamed2016implicit}. Utility-based evaluations assess how well predictive machine learning tasks or statistical analyses can be repeated on synthetic data, using measures of \emph{machine learning utility} or \emph{statistical utility} \cite{xu2019ctgan,fossing2024evaluationSynthcityModels,jordon2022synthetic}. So far, no gold standard has been established in measuring the quality of synthetic data. Good performance in one of the previous approaches does not imply the same for a different one \cite{theis2016note,zein2022foolXGB,jordon2022synthetic}. Moreover, none of these possibilities to measure synthetic data quality provides a data-based explanation of their scores or insights into \emph{why} the quality of some synthetic dataset has been evaluated to be good or bad.

\subsection{Explainable Artificial Intelligence}
\label{subsec:background_IML}

Due to their high performance, machine learning models are increasingly implemented in high-stake decision-making. However, the gain in performance often comes at the cost of reduced interpretability. XAI methods aim to enhance transparency and trustworthiness by developing inherently interpretable models or creating post-hoc methods that provide insights into the behavior of complex black box models.

Post-hoc XAI methods can be categorized along several dimensions, which should be taken into account when selecting the most suitable approach in a given context \cite{molnar2022interpretable,molnar2022pitfalls}:
\emph{Model-agnostic} methods can be applied to any model, whereas \emph{model-specific} methods are tailored to particular model classes, making them often more computationally efficient. For instance, Shapley additive explanation values (SHAP) \cite{lundberg2017SHAP} are not limited to any specific model class, DeepSHAP \cite{lundberg2017SHAP} and TreeSHAP \cite{lundberg2018treeSHAP} can only be used for deep learning architectures and tree-based learners, respectively.
Another central axis of differentiation is the scope of interpretability: \emph{Global} methods provide explanations for the overall model behavior across all data points, \emph{local} ones for single instances. Partial dependence plots (PDP) \cite{friedman2001PDP}, as an example, provide a global assessment of feature influence, while individual conditional expectation (ICE) \cite{Goldstein2015ICE} curves refine this perspective by depicting variation at a local level.
A further distinction arises regarding the perspective of explanations: \emph{Prediction-based} methods, such as SHAP, concentrate on how the model’s predictions change in response to different inputs, while \emph{loss-based} methods, such as permutation feature importance (PFI) \cite{Fisher2019PFI}, focus on how the model learns and generalizes rather than how it predicts. 
Explanation methods differ in how they account for relationships between features: \emph{Marginal} explanations evaluate the effect of a feature ignoring dependencies with other features while \emph{conditional} explanations account for feature relationships based on the data distribution \cite{blesch2025conditional,aas2019explaining,chen2023algorithms}. Marginal methods are said to be ``true to the model'' because they reflect how the model internally processes features, even if the resulting explanations are unrealistic given the data due to ignored correlations. In contrast, conditional methods are said to be ``true to the data'', ensuring that explanations align with real-world feature dependencies \cite{chen2020true}.

The reliability of any interpretability method highly depends on the model’s capacity to generalize. If a model underfits, explanations may reflect oversimplified patterns, whereas an overfitting model may yield explanations that are driven by noise rather than meaningful structure.

\section{Methods}
\label{sec:methods}

To not only assess the quality of synthetic data but also gain more granular insights into their limitations, we propose applying XAI methods to a synthetic data detection model. Specifically, we introduce methods capable of addressing questions \ref{enum:Q1} -- \ref{enum:Q4} from Section~\ref{sec:intro}, which represent increasing levels of explanatory detail.

In the following, $D \coloneqq \{D_\text{real}, D_\text{syn}\}$ denotes a dataset of size $n$ and dimensionality $p$ consisting of original data $D_\text{real}$ and equally-sized synthetic data $D_\text{syn}$ from an arbitrary (explicit or implicit) generative model $G$. Furthermore, let $\mathbf{x}_j$ denote the $j$-th feature of $D$, $\mathbf{x}_{-j}$ the subset of features excluding $j$, and $\mathbf{x}^{(i)}$ the $i$-th instance.

\subsection{Synthetic Data Detection Model}
\label{subsec:methods_detection}

A synthetic data detection model is a binary classifier $C: \mathcal{X} \rightarrow [0,1]$ trained on $X_\text{train} \subset D$ with corresponding binary labels $\mathbf{y}_\text{train}$ marking if an observation is real or synthetic. It outputs a probability $C(\mathbf{x})$ for a given data point to be real. Its performance on unseen test data $X_\text{test}$ and $\mathbf{y}_\text{test}$ can be evaluated by metrics such as accuracy and the area under the ROC curve (AUC), and serves as a measure for the quality of the synthetic data: Intuitively, high quality synthetic data should almost be indistinguishable from real data for $C$. False positive rates and false negative rates can be used to assess fidelity and diversity of the synthetic data \cite{simon2019revisiting}. If additionally a train-test-split is introduced before generative model training so that $C$ gets trained on different real data than $G$, the performance of $C$ also reflects the generalization ability of $G$.

Different types of binary classifiers can be used as a synthetic data detection model. In order to obtain reliable evaluation scores and explanations, it is vital to select a highly capable model and ensure it does not overfit substantially. For tabular data, well-tuned gradient-boosted tree ensembles such as XGBoost \cite{Chen2016XGBoost} are a robust choice \cite{zein2022foolXGB}.

\subsection{Explaining Synthetic Data Detection with XAI Methods}
\label{subsec:methods_IML}

Given a synthetic data classifier $C$ and a dataset of original and synthetic instances $D$, we present suitable XAI methods addressing questions \ref{enum:Q1} -- \ref{enum:Q4} and discuss their properties, advantages and shortcomings. For a more holistic and detailed discussion of XAI methods, we refer to Molnar \cite{molnar2022interpretable}.

\subsubsection{\ref{enum:Q1}: Feature Importance Measures}

To determine which features or feature combinations were not reproduced realistically in synthetic data, feature importance measures can be leveraged. High feature importance values can point to the features or dependencies most responsible for low synthetic data quality.

\emph{Permutation feature importance} (PFI) is a global loss-based method for assessing feature importance. It quantifies the impact of a feature $\mathbf{x}_j$ by the drop in model performance, measured by the model loss $L$, averaged over all instances when this feature is replaced by a permuted version $\mathbf{\tilde{x}}_j$ of itself:
\begin{equation*}
    \operatorname{PFI}_j = \frac{1}{n} \sum_{i = 1}^n \left( L\left(C(\mathbf{\tilde{x}}_j; \mathbf{x}_{-j}^{(i)}), y^{(i)}\right) - L\left(C(\mathbf{x}^{(i)}), y^{(i)}\right)\right).
\end{equation*}
As an effect of the permutation across the whole dataset, the associations between $\mathbf{x}_j$ and $y$, as well as all other features, are removed. 
The more the model's performance declines, the more important the feature is considered. Since the model's performance is involved in the computation of PFI, it is preferable to calculate it on test data \cite{molnar2022interpretable}. PFI can yield inaccurate results in the presence of highly correlated features as the permutation procedure leads to the generation of off-manifold data. This can be tackled by replacing the marginal permutation routine by conditional sampling techniques \cite{watson2021CPI,blesch2025conditional}.

Alternatively, Shapley values \cite{shapley1953SV} can be leveraged to obtain feature importances: The \emph{Shapley feature importance} \cite{lundberg2018treeSHAP} is  calculated as the mean absolute Shapley value
\begin{equation*}
 \phi_{j} = \frac{1}{n} \sum_{i = 1}^n \left|\phi_j^{i}\right|,
\end{equation*}
where $\phi_j^{i}$ denotes the Shapley value for instance $i$ and feature $j$. For an introduction to Shapley values, see Section~\ref{sec:methods_Shapley}.

\subsubsection{\ref{enum:Q2}: Feature Effect Plots}
The marginal effect of a feature on the detection model's prediction can be graphically represented by feature effect plots. This allows for the identification of unrealistic feature value regions in the synthetic data, as well as areas of the original data distribution that are not covered or underrepresented by the synthetic data. To facilitate understanding of feature effect plots in the context of synthetic data detection, Figure~\ref{fig:q2} and its interpretation in Section~\ref{sec:ex_q2} can already be considered alongside their methodological discussion in this section.

At a single-instance level, this can be visualized using \emph{individual conditional expectation} (ICE) curves, represented by the graph of the function
\begin{equation*}
    \operatorname{ICE}_j^i (x_j) = C(x_j; {\mathbf{x}_{-j}^{(i)}}),
\end{equation*}
where model predictions are evaluated for different values of the $j$-th feature while keeping all other features constant. Non-parallel ICE curves of different individuals indicate the presence of an interaction effect with other features on the prediction, suggesting that some feature dependencies present in the original data were not consistently retained in the synthetic data.

By averaging over each feature value, a global \emph{partial dependence plot} (PDP) can be derived from local ICE curves:
\begin{equation*}
    \operatorname{PDP}_j (x_j) = \frac{1}{n} \sum_{i=1}^{n} \operatorname{ICE}_j^i (x_j).
\end{equation*}
This provides a comprehensive view of the feature’s marginal effect on the prediction. Regions where the PDP drops significantly below 0.5 indicate unrealistic feature values in the synthetic data, making them easily identifiable for the classifier. Conversely, regions where the PDP rises well above 0.5 highlight feature value areas that are absent or underrepresented in the synthetic data. Often, PDP and ICE curves are presented together in a joint plot.

Both ICE and PDP assume independent features, which can lead to misleading interpretations when features are correlated. Accumulated local effects (ALE) \cite{apley2020ALE} plots are an alternative to PDP for this case, as they compute localized instead of marginal effects by measuring changes in model predictions within small conditional neighborhoods.

\subsubsection{\ref{enum:Q3}: Shapley Values}
\label{sec:methods_Shapley}

Rooted in cooperative game theory, \emph{Shapley values} \cite{shapley1953SV} were originally designed to fairly distribute a total payoff among players in a game. In the context of machine learning, the most common approach treats features as players, and distributes a single observation's prediction among them based on their contribution. This is done in an inclusion-removal manner considering all possible feature coalitions $S$: 
\begin{equation}
\label{eq:SV}
\phi_j^{i} = \sum_{S \subseteq \{1, \ldots, p\} \setminus \{j\}} \frac{|S|!(p-|S|-1)!}{p!} \left(v^i(S \cup \{j\}) - v^i(S)\right).
\end{equation}
The function $v^i$ returns the model’s expected output when only a given subset of features of an instance $i$ is considered. In the classical marginal approach \cite{lundberg2017SHAP}, the remaining features are marginalized out using their \textit{marginal} expectations, implicitly assuming independence between observed and unobserved features. In the conditional approach \cite{aas2019explaining}, the remaining features are integrated out based on their \textit{conditional} expectations given the considered features, preserving dependencies in the data. This results in conditional Shapley values, which are often more realistic when estimated accurately. However, their estimation can be challenging and entails higher computational costs \cite{aas2019explaining,chen2023algorithms}. Equation~\eqref{eq:SV} can be slightly modified to compute contributions of feature interactions \cite{sundararajan2020taylorInteract}.
For large datasets and high-dimensional feature spaces, computing exact Shapley values can be computationally expensive or even infeasible. Often, only a subset of all possible coalitions is considered to make computations feasible, for instance with the model-agnostic KernelSHAP method \cite{aas2019explaining,lundberg2017SHAP}. Also model-specific adaptions such as DeepSHAP and TreeSHAP can speed up calculations significantly.

In the setting of synthetic data detection, Shapley values decompose the predictions of individual synthetic observations to reveal which features (or feature interactions, in the case of Shapley interaction values) make instances appear unrealistic to the detection model. Conversely, they also help identify which values and feature combinations of a real instance are insufficiently represented in the synthetic data distribution.
Thus, Shapley values quantify the relevance of the \emph{presence} of each feature value (or feature combination) in determining how realistic this observation appears to the detection model:
For real observations with $C(\mathbf{x})\gg 0.5$, the features with the largest Shapley values indicate that the feature values they correspond to are the main reasons the detection model is able to separate them from the synthetic data. This suggests that these feature values may be underrepresented in the synthetic data. The same applies to synthetic observations with $C(\mathbf{x}) \ll 0.5$.

\subsubsection{\ref{enum:Q4}: Counterfactuals}

Counterfactual explanations (CE) present examples of minimal feature changes that would alter the model's prediction to a different outcome.
For our binary classification setting, a different outcome will correspond to crossing the threshold of 0.5. 
We will leverage this method to study how we can most easily modify the feature values of a correctly detected synthetic sample so that it looks realistic to the detection model (i.e., has $C(\mathbf{x}) > 0.5$). 
The best counterfactual examples possess several quality properties \cite{guidotti2024counterfactual}, where sparsity/proximity (changes are as few and small as possible), plausibility (changes are realistic and align with the feature distribution), validity (actually resulting in a changed outcome) are the most relevant in our case.

As discussed in Section~\ref{sec:related-work}, many methods for counterfactual generation rely on generative models themselves in order to produce realistic counterfactuals. The \emph{Monte Carlo sampling of realistic counterfactual explanations} (MCCE) method \cite{redelmeier2024mcce}, used in Section~\ref{sec:ex_q4}, directly aims to achieve the aforementioned quality properties and leverages an autoregressive tree-based synthesizer to generate on-manifold counterfactuals, making it particularly well-suited for tabular data.

\section{Real Data Examples}
\label{sec:examples}

We demonstrate our approach on real world data. First, we examine the performance of different synthetic data detection models (Section~\ref{sec:ex_peform}) on various datasets and generative models. After that, we answer questions \ref{enum:Q1} -- \ref{enum:Q4} with the XAI methods presented in Section~\ref{subsec:methods_IML} for synthetic adult data \cite{adult_data} generated with TabSyn \cite{zhangTabSyn} (Section~\ref{sec:ex_q1-4}). In Appendix~\ref{ap:nursery}, we do a similar, but simplified exercise for synthetic nursery data \cite{nursery_data} generated with CTGAN \cite{xu2019ctgan}. Generic scripts for detection model tuning and training, for applying the XAI methods discussed in this paper, as well as for reproducing our presented tables and figures can be found in the repository accompanying this paper.\footnote{\url{https://github.com/bips-hb/XAI_syn_data_detection}.} Additionally, it includes an overview and descriptive statistics of all datasets and further context for our experiments with all synthesizers considered.

\subsection{Performance of Different Synthetic Data Detection Models}
\label{sec:ex_peform}

\begin{figure}[t]
    \centering
    \includegraphics[width=\linewidth]{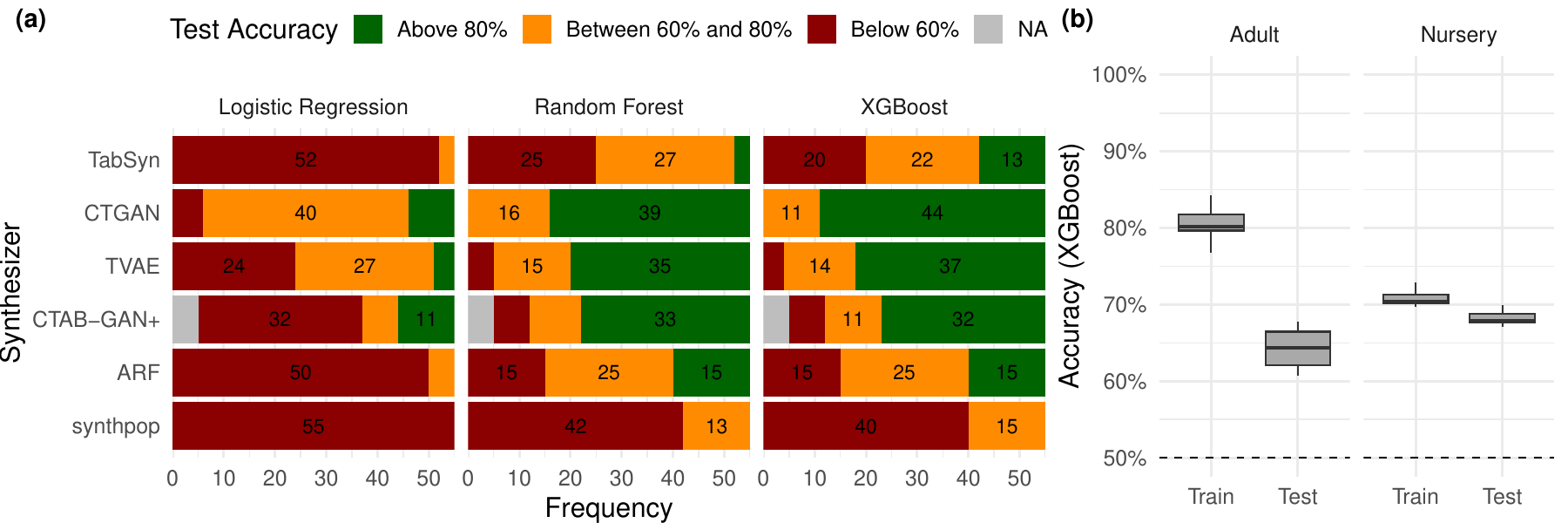}
    \caption{(a) Synthetic data detection performance of logistic regression, random forest \cite{breiman2001RF} and XGBoost models with six generative models generating five synthetic datasets for eleven original datasets each. CTAB-GAN+ did not converge for all runs. (b) Synthetic data detection performance for XGBoost on train and test data for adult and nursery data with ten replications.}
    \label{fig:detection_performance}
\end{figure}

In order to obtain reliable explanations, we need to ensure that we use a high performing classifier as our synthetic data detection model. For this purpose, we performed hyperparameter tuning for XGBoost via Bayesian optimization, e.g., for the tree depth, learning rate and regularization parameters. As already stated in \ref{subsec:methods_detection}, a well-tuned XGBoost model is typically a robust choice for classification tasks on tabular data. Figure~\ref{fig:detection_performance}a supports this statement: Across six different state-of-the-art or frequently used generative models \cite{zhangTabSyn,xu2019ctgan,watson2023adversarial,nowok2016synthpop,zhao2024ctab} and eleven standard machine learning datasets from publicly available sources \cite{UCI_Repo,kaggle_repo}, tuned XGBoost models consistently show the strongest classification performance against logistic regression and random forest as baseline models.

Figure~\ref{fig:detection_performance}b shows the performance of the tuned XGBoost models for the two datasets (adult and nursery) we use in the following to illustrate our approach.

\subsection{Answering \ref{enum:Q1} -- \ref{enum:Q4} for Synthetic Adult Data}
\label{sec:ex_q1-4}

Before synthesis, we removed all rows with missing values from the adult dataset as well as the categorical feature \texttt{education}, since its information value is identical to \texttt{education\_num}, resulting in a dataset with 47\,876 instances and 14 columns (6 numeric and 8 categorical). We generated ten synthetic datasets using TabSyn, a state-of-the-art DDPM for tabular data. On each of these generated datasets, we trained an XGBoost model for synthetic data detection using Bayesian optimization for hyperparameter tuning and a train-test split of 30\%, while keeping the real data points the same across all ten synthetic datasets.

\subsubsection{\ref{enum:Q1}}

\begin{figure}[t]
    \centering
    \includegraphics[width=1\linewidth]{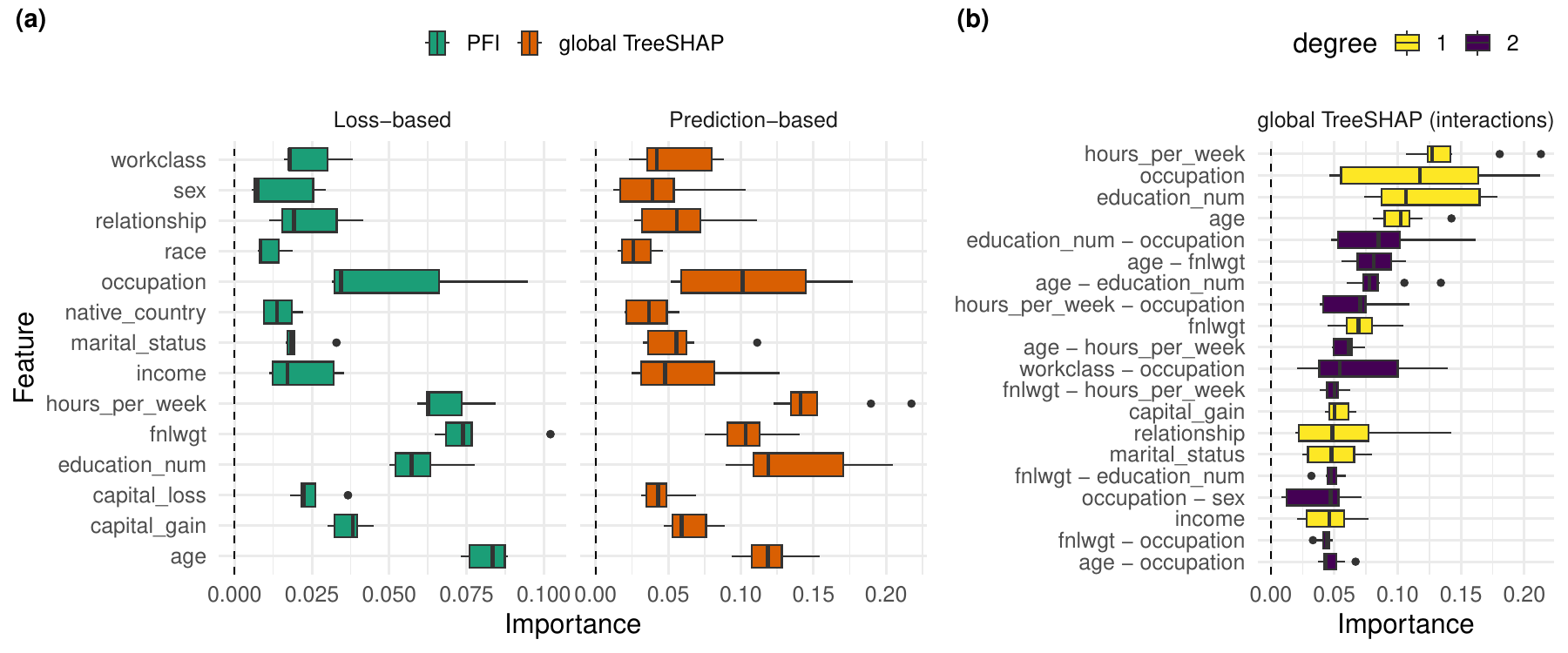}
    \caption{Feature importance values for synthetic data detection with XGBoost for ten TabSyn-generated synthetic adult datasets. Higher importance values indicate poorer synthesis quality. (a) PFI and global TreeSHAP values. (b) Global TreeSHAP interaction values of degree 1 and 2 (top 20 most important).}
    \label{fig:q1}
\end{figure}

To address our first research question on identifying the most challenging features and feature dependencies for TabSyn, we analyze feature importances using both loss-based and prediction-based XAI methods. Figure~\ref{fig:q1}a shows results for PFI and global TreeSHAP (mean absolute Shapley values estimated by TreeSHAP), while Figure~\ref{fig:q1}b provides a finer examination of the interaction effects inherently obtained using TreeSHAP. In general, if a feature has a high importance, it indicates patterns in the synthetic data that make a differentiation on a global level easier for the detection model, whereas low importance suggests more realistic feature values.

At a broad level, both methods in Figure~\ref{fig:q1}a identify similar key features for the XGBoost detection model, such as \texttt{age}, \texttt{hours\_per\_week}, \texttt{fnlwgt}, and \texttt{education\_num}, highlighting these features as weak spots in the synthetic data where marginal distributions or dependencies to other features are not replicated accurately. Moreover, the relatively low variance (except for \texttt{occupation}) in the boxplots suggests that the detection models rely on the same features across all ten synthesized dataset versions, indicating a consistent decision basis. The high variance for \texttt{occupation} may be due to its strong dependencies with multiple important features (see Figure~\ref{fig:q1}b), which seem challenging to retain for TabSyn. Comparing PFI and global TreeSHAP, both yield similar feature rankings and relative differences. However, their importance values differ in scale: As mentioned in \ref{subsec:methods_IML}, TreeSHAP sums absolute contributions across instances, leading to larger values, while PFI measures the average performance drop of a feature removal. An exception is the feature \texttt{fnlwgt}, which is ranked considerably higher in PFI than in TreeSHAP. This is likely due to its strong interactions with other important features, such as \texttt{age} and \texttt{hours\_per\_week} (see Figure~\ref{fig:q1}b). Since PFI disrupts feature dependencies through permutation, it fully attributes interaction effects to each involved feature once, not multiple times, whereas TreeSHAP distributes them fairly among the interacting features \cite{fumagalli2025unifying}.

Figure~\ref{fig:q1}b shows that, alongside strong main effects, many interaction effects contribute substantially to the model’s predictions, which are inherently calculated with TreeSHAP for XGBoost models (interaction TreeSHAP). The features \texttt{hours\_per\_week}, \texttt{education\_num}, and \texttt{age} remain the most important individually, but interactions, such as \texttt{education\_num} with \texttt{occupation} and \texttt{age} with \texttt{fnlwgt} and \texttt{education\_num}, also show a strong influential effect. Their appearance within the most important features (main and interactions) indicates that TabSyn struggles to fully recreate this complex dependence structure.
An interesting case is \texttt{occupation}, which, as previously noted, appears in numerous interactions (with \texttt{education\_num}, \texttt{hours\_per\_week} and \texttt{workclass}). The variability in this interaction across all ten dataset versions further confirms that the dependencies involving \texttt{occupation} are difficult for the synthesizer to capture consistently.
Another notable observation is that the importance ranking of \texttt{occupation} in Figure~\ref{fig:q1}b including the interactions is much higher than in both PFI and TreeSHAP in Figure~\ref{fig:q1}a. One possible explanation is that \texttt{occupation}’s interactions partially cancel out or reduce the main effect due to different contribution signs in the standard TreeSHAP decomposition, whereas for interaction TreeSHAP, the separation of interactions prevents this effect. This may happen if the marginal distribution of \texttt{occupation} deviates from the real distribution (i.e., making it easier to distinguish), but the dependencies for influential classes of \texttt{occupation} to other features are realistically generated.

\subsubsection{\ref{enum:Q2}}
\label{sec:ex_q2}

\begin{figure}[t]
    \centering
    \includegraphics[width=\linewidth]{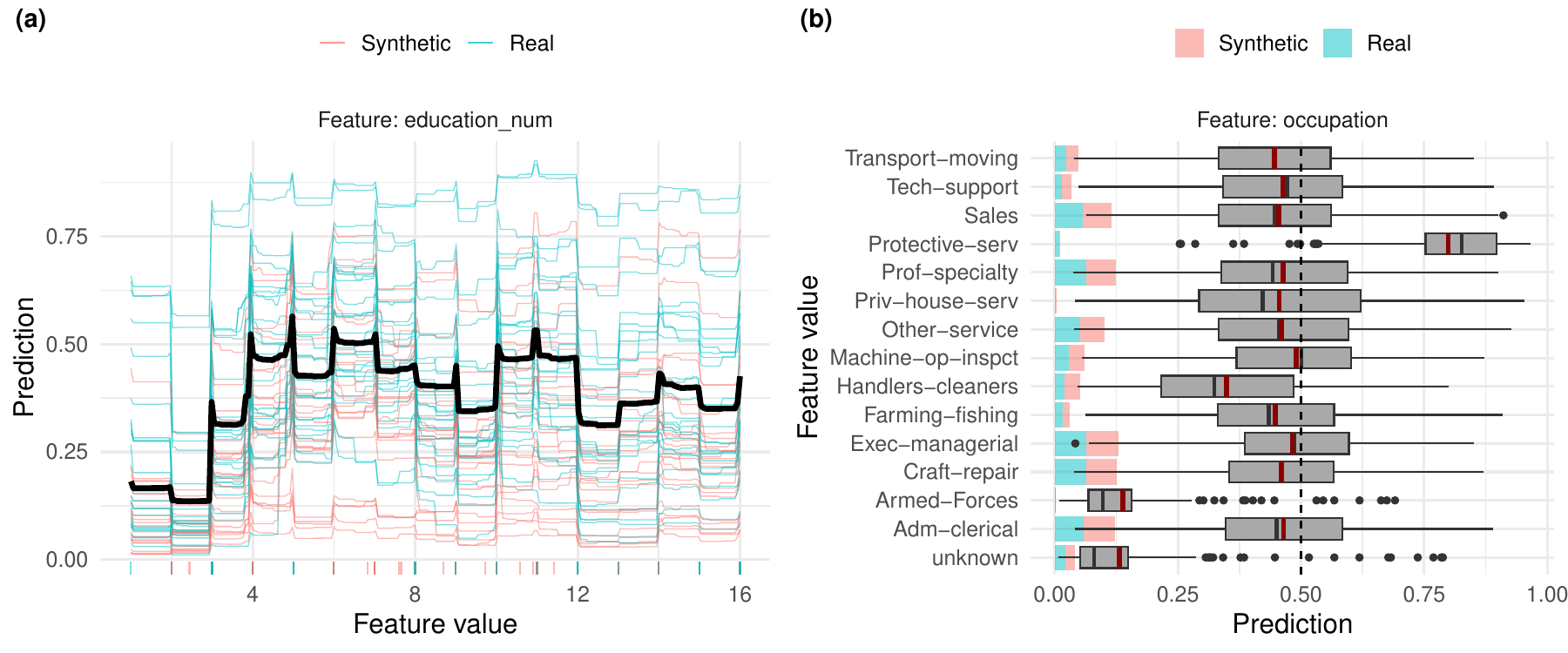}
    \caption{ICE/PDP for synthetic data detection with XGBoost for TabSyn-generated synthetic adult data. (a) Numeric feature \texttt{education\_num}, distribution for original and synthetic data on x-axis. (b) Categorical feature \texttt{occupation}, PDP in red, frequencies for real and synthetic data on y-axis.}
    \label{fig:q2}
\end{figure}

Using feature effect plots, we now take a closer look at the numerical feature \texttt{education\_num} and the categorical feature \texttt{occupation} which were among the top-ranked features and appeared in several interactions in Figure~\ref{fig:q1}b. From now on, we base our explanations on one synthetic dataset only and do not further examine the variation across different TabSyn-generated datasets.
ICE curves and PDP for \texttt{education\_num} are shown in Figure~\ref{fig:q2}a with indicated marginal distributions for real and synthetic data on the x-axis. We observe that especially in the low-density region of the real data for feature values below 4, the PDP is located considerably below 0.5, implying that TabSyn learned unrealistic patterns in this area which make the identification of synthetic data easy for XGBoost. Furthermore, the PDP and the plotted data distributions reveal unrealistic synthetic feature values: While in the real data only integer values occur for \texttt{education\_num}, TabSyn also generates non-integer values, resulting in a periodical PDP drop between integers. Looking at the ICE curves, we see that not all curves run in parallel and that some cross each other. This indicates the presence of interactions with \texttt{education\_num} which alter the course of the curves individually on top of the marginal effects, which aligns to the findings presented in Figure~\ref{fig:q1}b. Another observation is that the share of real data ICE curves markedly falling under 0.5 is higher than the share of synthetic data ICE curves over 0.5. This means that it is harder for the model to correctly classify real data than synthetic data based on this feature, indicating low fidelity but high diversity of the synthetic data for this feature.

Figure~\ref{fig:q2}b shows results for the categorical variable \texttt{occupation}. For each class, the variation of ICE values across instances is displayed by box plots, the PDP values are represented by red lines. On the y-axis, the class frequencies are shown for real and synthetic data. We observe that the classes \texttt{Protective-serv}, \texttt{Armed-Forces} and \texttt{unknwon} with most deviation from 0.5 have a low number of occurrences, while the most frequent classes such as \texttt{Prof-specialty} and \texttt{Exec-managerial} tend to have values close to 0.5. This implies that data generation quality is low for rare classes and improves for more frequent ones. Looking at \texttt{Protective\_serv} and the corresponding class frequencies for real and synthetic data, we notice that this class is substantially underrepresented in the synthetic data, which can be an explanation for its PDP value close to 1 (prediction as real). On the contrary, the class \texttt{Handlers-cleaners} seems to be moderately overrepresented in the synthetic data, which can be the reason why its PDP and ICE values are consistently smaller than 0.5 (prediction as synthetic). The class \texttt{unknown} appears to be equally represented in real and synthetic data. However, its PDP value is close to 0, indicating that synthetic instances with this class are often unrealistic. A reason for this can be that combinations of this class with other feature values are not retained realistically in the synthetic data. This aligns to the presence of important interactions with \texttt{occupation} in \ref{fig:q1}b.

\subsubsection{\ref{enum:Q3}}
\label{sec:ex_Q3}


While the first two questions focused mainly on the detection model's global behavior, we now examine how features and feature interactions contribute to the classification of specific individual predictions.
To explore this, we analyze two correctly classified samples -- one synthetic and one real. 
Note that these two examples illustrate key aspects that should be considered when interpreting predictions of the detection model. 
However, for a comprehensive assessment of local behavior, a structured analysis across many more observations is required.

\begin{figure}[t]
    \centering
    \includegraphics[width=1\linewidth]{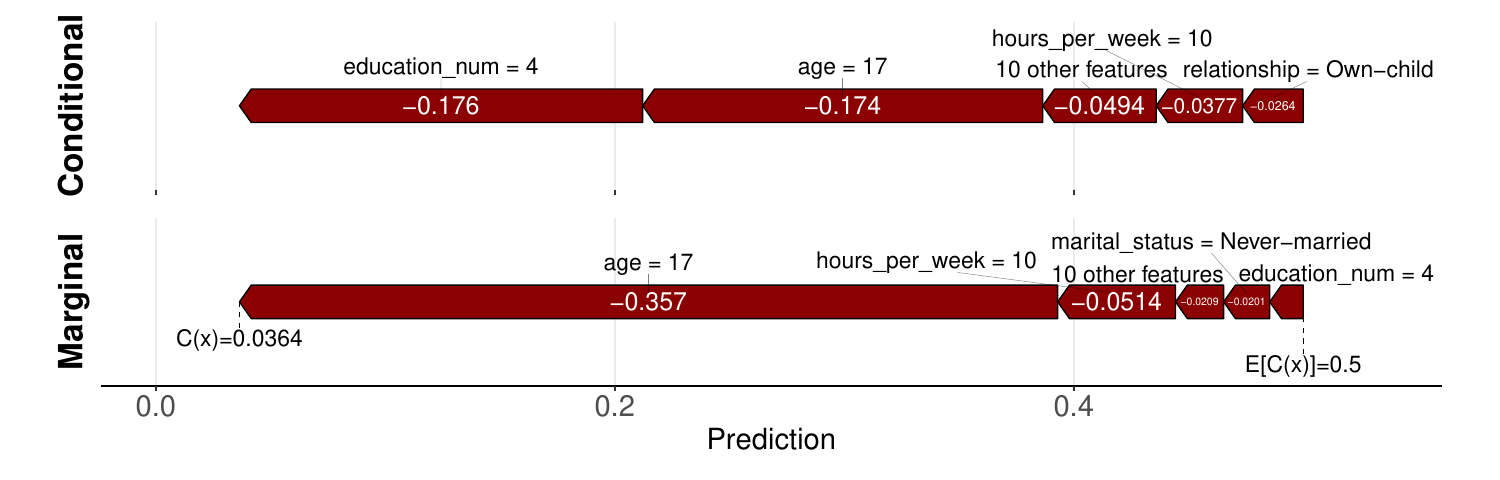}
    \caption{Force plots for conditional and marginal Shapley values decomposing the XGBoost prediction for an exemplary instance of Tabsyn-generated synthetic adult data.} 
    \label{fig:q3_cond_vs_marg_syn}
\end{figure}

We first explore a specific synthetic observation corresponding to a young (\texttt{age} $ =17$), white female with low education (\texttt{education\_num} $ =4$, indicating education up to the 7th or 8th grade) based in the United States with marginal and conditional Shapley values. Note that potential issues with the region around \texttt{education\_num} $  =4$ were already visible in Figure~\ref{fig:q2}a. The feature distributions in the conditional approach are estimated with conditional inference trees \cite{redelmeier2020} and both approaches use the KernelSHAP method with 2000 coalitions to approximate the Shapley values. The force plots in Figure~\ref{fig:q3_cond_vs_marg_syn} visualize how differently marginal and conditional Shapley values drive the prediction from the baseline value of 0.5 down to a very low predicted probability of $C(\mathbf{x})=0.0364$: The marginal approach attributes the majority of the contribution to this prediction to the young age, whereas the conditional approach distributes the contribution almost evenly between \texttt{education\_num} and \texttt{age}. As noted already in Section \ref{subsec:methods_IML}, the marginal and conditional approaches differ in the presence of (local) feature dependence. Upon closer inspection of these two features, we find that the combination present in our synthetic example is highly unlikely in the real data: Only 0.7\% of 17-years-old real individuals have a value of 4 for \texttt{education\_num}, the average value is 6.69 for this age. Moreover, the mean age for \texttt{education\_num} $  =4$ is the oldest across all education levels with 49.12 years. An explanation for these numbers can be the fact that it was common for older generations to quit school after the 8th grade to start working, as younger generations typically attend school for longer than that. 
Moreover, their global correlation in the real data is small (0.03), but larger for the younger (0.54 for \texttt{age}$\leq 20$).
The stronger feature dependence among younger individuals is natural, as many are still in education and have yet to attain higher degrees.
With this in mind, we consider the conditional approach the more appropriate, and argue that this synthetic sample is weak not only in terms of \texttt{age}, but also in \texttt{education\_num}.

\begin{figure}[t]
    \centering
    \includegraphics[width=1\linewidth]{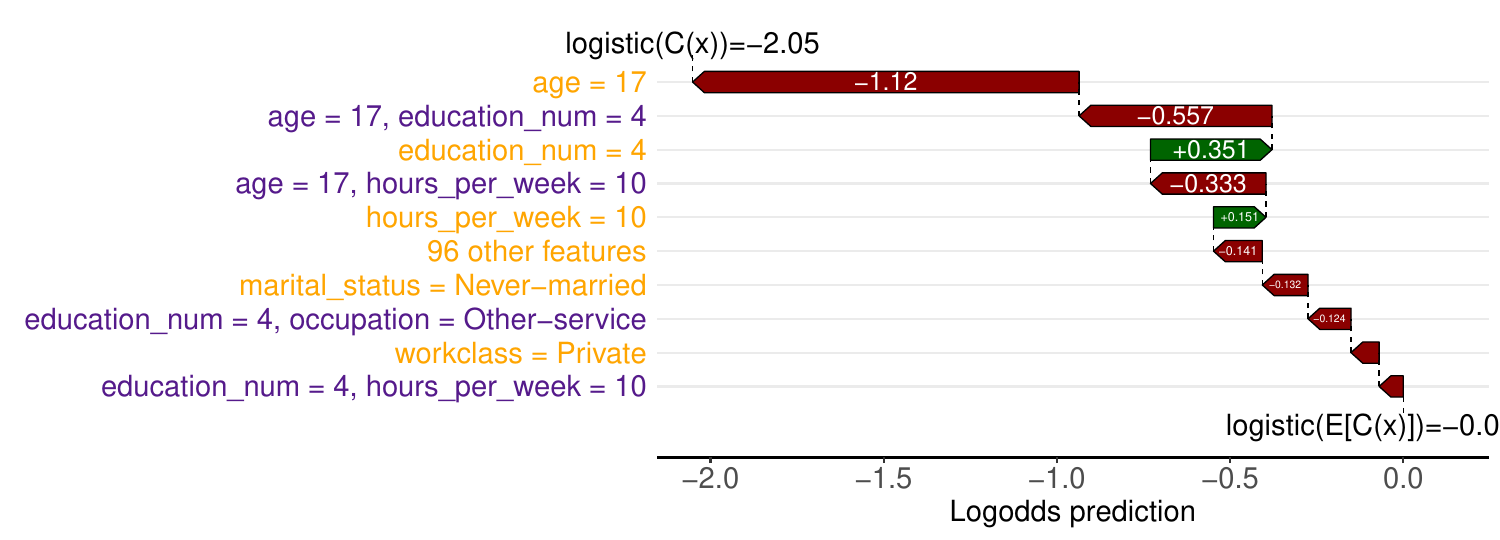}
    \caption{Waterfall plots for Shapley interaction values decomposing the XGBoost prediction for an examplary instance of TabSyn-generated synthetic adult data. Note that TreeSHAP decomposes the prediction on the log-odds/logistic scale rather than on the probability scale.}
    \label{fig:q3_inter_syn}
\end{figure}

Figure~\ref{fig:q3_inter_syn} confirms this line of reasoning: It displays a waterfall plot with the largest (in absolute value) Shapley interaction values for the same synthetic female as above, computed with the (path dependent) TreeSHAP algorithm \cite{lundberg2018treeSHAP}, which partially accounts for feature dependencies, placing it between the conditional and marginal approaches:
The interaction between \texttt{age} and \texttt{education\_num} is identified as a key factor for correctly classifying the sample. Interestingly, when adjusting for interactions, \texttt{education\_num} alone actually makes the sample look more similar to the real observations. This suggests that locally around this sample, \texttt{education\_num} itself aligns well with real observations, but its combination with the specific \texttt{age} is causing the discrepancy.

\begin{figure}[t]
    \centering
    \includegraphics[width=1\linewidth]{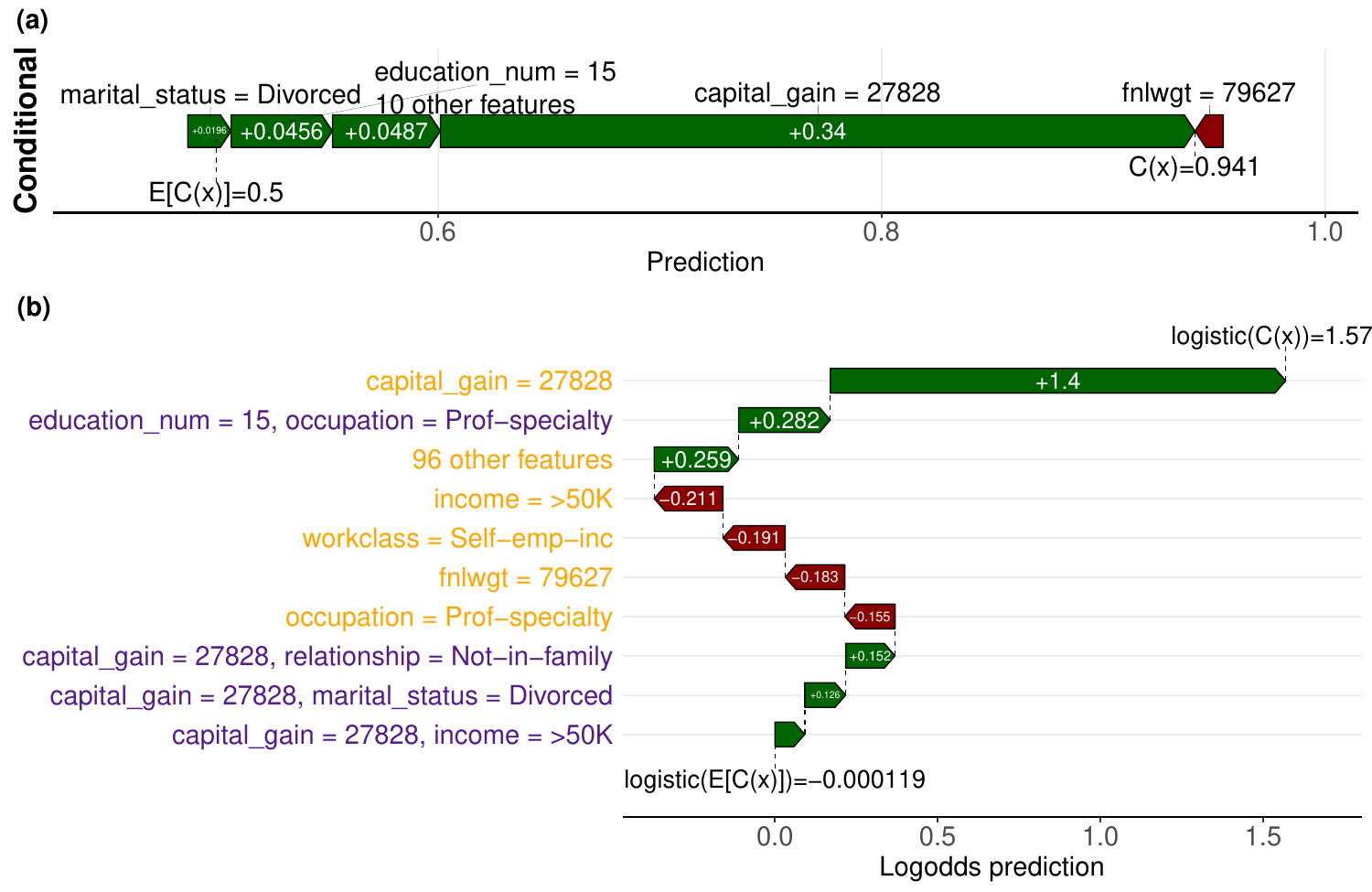}
    \caption{Force and waterfall plots for, respectively, (a) conditional Shapley values and (b) Shapley interaction values decomposing the XGBoost prediction for an examplary real instance of the adult data. Note that TreeSHAP decomposes the prediction on the log-odds/logistic scale rather than on the probability scale.}
    \label{fig:q3_cond_inter_real}
\end{figure}

For the majority of the features, the overall average dependence in this dataset is relatively small, such that the conditional and marginal Shapley values more or less agree for most of the samples. However, as seen above, the features may be highly dependent locally. Going forward, we therefore stick to the conditional approach. 
Figure~\ref{fig:q3_cond_inter_real} displays both conditional Shapley values and Shapley interaction values for a 47-year-old, self-employed, male in the real dataset, which was easily correctly classified by the detection model $C(\mathbf{x})=0.94$. The main contributor to that was the \texttt{capital\_gain} feature, while the interactions also reveal that the dependencies between \texttt{education\_num} and \texttt{occupation} play a role.
As seen from Figure~\ref{fig:q1}, \texttt{capital\_gain} was not very important globally, which exemplifies that there may be local areas in the feature space which are not adequately represented in the synthetic data even if the feature is globally well represented. The \texttt{education\_num}--\texttt{occupation} interaction, on the other hand, is already highlighted as a potential issue globally.

\subsubsection{\ref{enum:Q4}}
\label{sec:ex_q4}

\begin{figure}[t]
    \centering
    \includegraphics[width=1\linewidth]{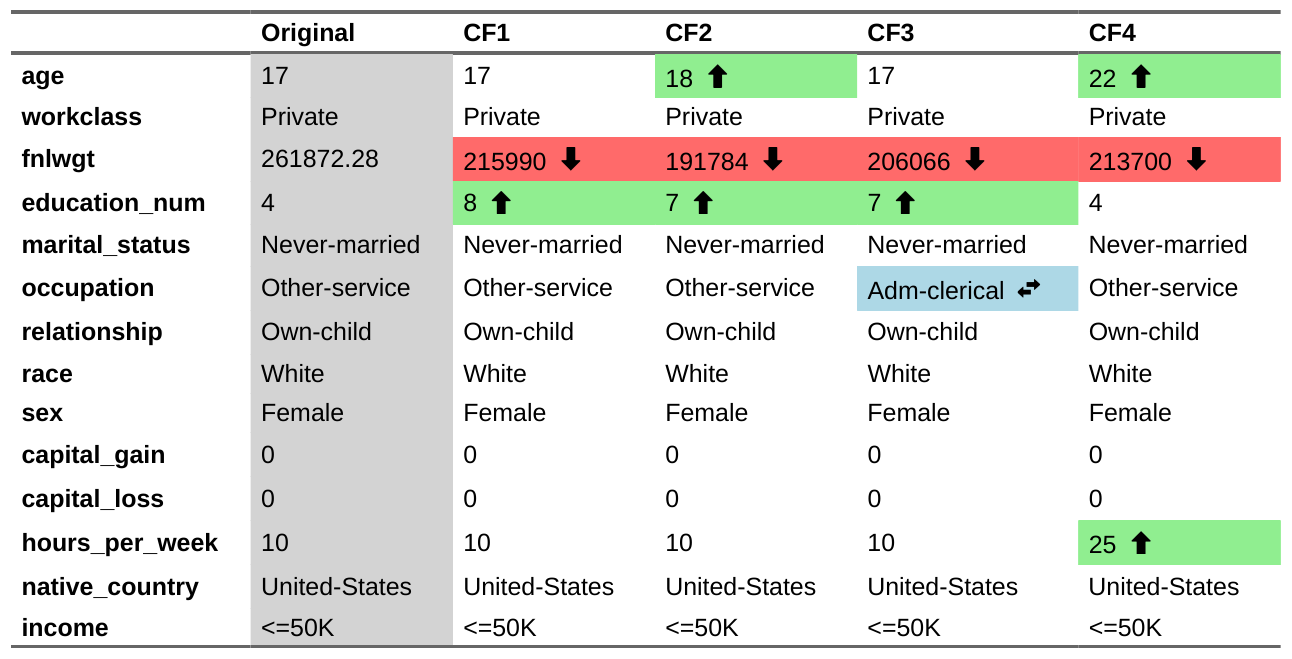}
    \caption{Four counterfactual explanations for an examplary instance of TabSyn-generated synthetic adult data. Highlighted features are changed.}
    \label{fig:q4}
\end{figure}

Finally, we investigate how minimal changes to synthetic observations can be performed to make them look realistic. We proceed to exemplify this with the same synthetic sample as in Section~\ref{sec:ex_Q3}, the young female with low education. Figure~\ref{fig:q4} provides four counterfactual examples which modify a few of the feature values in different ways to change the classification to real. The counterfactual examples are generated with the MCCE method of \cite{redelmeier2024mcce} with all features mutable and $5\times10^5$ Monte Carlo samples. 

First, observe that \texttt{fnlwgt} is changed in all counterfactuals. This feature reflects the number of people each observation represents in the population, making it natural that it must be adjusted when other features change to maintain consistency. Moreover, Figure~\ref{fig:q1}b indicated issues in the synthetic data for its dependencies with other features on a global level, so that its value might not have aligned to other features. Apart from \texttt{fnlwgt}, the first counterfactual (CF1) changes only \texttt{education\_num}, while CF2 and CF3 perform slightly smaller changes in
\texttt{education\_num} while also increasing respectively \texttt{age} by 1 year and \texttt{occupation}.
Finally, CF4 illustrates that \texttt{education\_num} does not need to be changed for the synthetic observation to appear real -- increasing \texttt{age} by five years and \texttt{hours\_per\_week} by 15 hours also changes the classification to real.

\section{Discussion}

In this paper, we demonstrate that the application of XAI methods on a synthetic data detection model can generate valuable insights about synthetic data quality. We provide a set of suitable global and local XAI tools, such as feature importance measures, feature effect plots, Shapley values and counterfactuals. These XAI tools can be used to identify the most challenging features and dependencies in the original data for successful data synthesis and to detect unrealistic patterns in the synthetic data on dataset- and single-instance level. The experiments in Section~\ref{sec:examples} illustrate this on real world data and underline the gain in explanatory depth in comparison to traditional synthetic data quality metrics and visual approaches such as histograms and correlation plots. The insights generated from our suggested XAI-driven analysis can be used for synthetic data auditing to explain overall utility, synthesis quality in low density areas of the real data, and to determine low fidelity instances. It could also be leveraged to analyze and compare the strengths and weaknesses of different generative models on different datasets and to debug or further improve generative models.

One limitation of this approach is that it is not directly applicable to a generative model itself but depends on the performance of a binary classifier used for synthetic data detection. Special care has to be taken during the tuning and training process of this classifier as low detection accuracy values can both be caused by insufficient classification performance and by high quality synthetic data. For some purposes, the detection model can even be too sensitive and might detect deviations such as different numeric precision which might not be relevant to assess the practical utility of the synthetic data. Moreover, it is critical to choose an appropriate XAI method considering data dependencies and explanation goal in order to obtain reliable explanations. As is generally true for all applications of XAI methods, there is no one-fits-all solution and it is important to know the strength and limitations of different tools. For instance, it is important to be aware of the differences of loss-based and prediction-based methods as well as marginal and conditional approaches \cite{molnar2022pitfalls}. Some of the XAI methods, especially conditional approaches and counterfactuals generators, are even based on generative modeling themselves. As stated in Section~\ref{subsec:background_genmodel}, one should keep in mind that generative models for tabular data synthesis are not yet as mature as for image and text data generation, which could lower the quality of explanations created with these methods.

We restricted this work to the explanation of synthetic data fidelity and diversity and did not consider the generalization and privacy dimension of data quality. Following a different routine for generative model training as described in Section~\ref{subsec:methods_detection}, our approach could be extended to cover these aspects as well. We leave this for future work.

Especially in the light of the current worldwide rise of generative models and the increasing use of synthetic data in all relevant areas of daily life, we strongly believe that the identification and explainability of synthetic data quality as well as the intersection of XAI and generative AI in general is of highest importance.

\label{sec:conclusion}

\section*{Acknowledgements}
This project was supported by the German Research Foundation (DFG), Grant Numbers 437611051, 459360854, by the U Bremen Research Alliance/AI Center for Health Care, financially supported by the Federal State of Bremen, the Norwegian Research Council through the Integreat Center of Excellence, Project Number 332645, and EU's HORIZON Research and Innovation Programme, project ENFIELD, grant number 101120657.
The project was initiated at a workshop supported by a PhD grant of the Minds, Media, Machines Integrated Graduate School Bremen.

We thank Marvin N. Wright for his valuable comments during the revision of the first draft of the manuscript.

This version of the contribution has been accepted after peer review for publication in the proceedings after the Third World Conference on eXplainable Artificial Intelligence, XAI-2025, but is not the version of record and does not reflect post-acceptance improvements, or any corrections. A link to the version of record will be included here upon publication. 

\bibliographystyle{unsrt}
\bibliography{main.bib}

\begin{thebibliography}{10}

\bibitem{vaswani2017attention}
Ashish Vaswani, Noam Shazeer, Niki Parmar, Jakob Uszkoreit, Llion Jones, Aidan~N Gomez, \L~ukasz Kaiser, and Illia Polosukhin.
\newblock Attention is all you need.
\newblock In {\em Advances in Neural Information Processing Systems}, volume~30, 2017.

\bibitem{ho2020denoising}
Jonathan Ho, Ajay Jain, and Pieter Abbeel.
\newblock Denoising diffusion probabilistic models.
\newblock In {\em Advances in Neural Information Processing Systems}, volume~33, pages 6840--6851, 2020.

\bibitem{song2020score}
Yang Song, Jascha Sohl-Dickstein, Diederik~P Kingma, Abhishek Kumar, Stefano Ermon, and Ben Poole.
\newblock Score-based generative modeling through stochastic differential equations.
\newblock In {\em International Conference on Learning Representations}, 2021.

\bibitem{jordon2022synthetic}
James Jordon, Lukasz Szpruch, Florimond Houssiau, Mirko Bottarelli, Giovanni Cherubin, Carsten Maple, Samuel~N Cohen, and Adrian Weller.
\newblock Synthetic data--what, why and how?
\newblock {\em arXiv preprint arXiv:2205.03257}, 2022.

\bibitem{lopez2022revisiting}
David Lopez-Paz and Maxime Oquab.
\newblock Revisiting classifier two-sample tests.
\newblock In {\em International Conference on Learning Representations}, 2017.

\bibitem{alaa2022faithful}
Ahmed Alaa, Boris Van~Breugel, Evgeny~S Saveliev, and Mihaela van~der Schaar.
\newblock How faithful is your synthetic data? {S}ample-level metrics for evaluating and auditing generative models.
\newblock In {\em International Conference on Machine Learning}, volume 162 of {\em PMLR}, pages 290--306, 2022.

\bibitem{schneider2024explainable}
Johannes Schneider.
\newblock Explainable generative {Ai} ({GenXAI}): A survey, conceptualization, and research agenda.
\newblock {\em Artificial Intelligence Review}, 57(11):289, 2024.

\bibitem{higgins2017beta}
Irina Higgins, Loic Matthey, Arka Pal, Christopher Burgess, Xavier Glorot, Matthew Botvinick, Shakir Mohamed, and Alexander Lerchner.
\newblock beta-{VAE}: Learning basic visual concepts with a constrained variational framework.
\newblock In {\em International Conference on Learning Representations}, 2017.

\bibitem{chen2016infogan}
Xi~Chen, Yan Duan, Rein Houthooft, John Schulman, Ilya Sutskever, and Pieter Abbeel.
\newblock {InfoGAN}: Interpretable representation learning by information maximizing generative adversarial nets.
\newblock In {\em Advances in Neural Information Processing Systems}, volume~29, 2016.

\bibitem{dombrowski2024trade}
Mischa Dombrowski, Hadrien Reynaud, Johanna~P M{\"u}ller, Matthew Baugh, and Bernhard Kainz.
\newblock Trade-offs in fine-tuned diffusion models between accuracy and interpretability.
\newblock {\em Proceedings of the AAAI Conference on Artificial Intelligence}, 38(19):21037--21045, 2024.

\bibitem{abnar2020quantifying}
Samira Abnar and Willem Zuidema.
\newblock Quantifying attention flow in transformers.
\newblock In {\em Proceedings of the 58th Annual Meeting of the Association for Computational Linguistics}, pages 4190--4197, 2020.

\bibitem{vig2019multiscale}
Jesse Vig.
\newblock A multiscale visualization of attention in the transformer model.
\newblock In {\em Proceedings of the 57th Annual Meeting of the Association for Computational Linguistics: System Demonstrations}, pages 37--42, 2019.

\bibitem{nagisetty2020xai}
Vineel Nagisetty, Laura Graves, Joseph Scott, and Vijay Ganesh.
\newblock {xAI-GAN}: Enhancing generative adversarial networks via explainable {AI} systems.
\newblock {\em arXiv preprint arXiv:2002.10438}, 2020.

\bibitem{zhou2018activation}
Zhiming Zhou, Han Cai, Shu Rong, Yuxuan Song, Kan Ren, Weinan Zhang, Jun Wang, and Yong Yu.
\newblock Activation maximization generative adversarial nets.
\newblock In {\em International Conference on Learning Representations}, 2018.

\bibitem{nguyen2016synthesizing}
Anh Nguyen, Alexey Dosovitskiy, Jason Yosinski, Thomas Brox, and Jeff Clune.
\newblock Synthesizing the preferred inputs for neurons in neural networks via deep generator networks.
\newblock In {\em Advances in Neural Information Processing Systems}, volume~29, 2016.

\bibitem{dravid2022medxgan}
Amil Dravid, Florian Schiffers, Boqing Gong, and Aggelos~K. Katsaggelos.
\newblock {medXGAN}: Visual explanations for medical classifiers through a generative latent space.
\newblock In {\em 2022 IEEE/CVF Conference on Computer Vision and Pattern Recognition Workshops (CVPRW)}, pages 2935--2944, 2022.

\bibitem{blesch2025conditional}
Kristin Blesch, Niklas Koenen, Jan Kapar, Pegah Golchian, Lukas Burk, Markus Loecher, and Marvin~N Wright.
\newblock Conditional feature importance with generative modeling using adversarial random forests.
\newblock {\em arXiv preprint arXiv:2501.11178}, 2025.

\bibitem{agarwal2020explaining}
Chirag Agarwal and Anh Nguyen.
\newblock Explaining image classifiers by removing input features using generative models.
\newblock In {\em Computer Vision -- ACCV 2020}, pages 101--118, 2020.

\bibitem{nemirovsky2022countergan}
Daniel Nemirovsky, Nicolas Thiebaut, Ye~Xu, and Abhishek Gupta.
\newblock {CounteRGAN}: Generating counterfactuals for real-time recourse and interpretability using residual gans.
\newblock In {\em Proceedings of the Thirty-Eighth Conference on Uncertainty in Artificial Intelligence}, volume 180 of {\em PMLR}, pages 1488--1497, 2022.

\bibitem{dandl2024countarfactuals}
Susanne Dandl, Kristin Blesch, Timo Freiesleben, Gunnar K{\"o}nig, Jan Kapar, Bernd Bischl, and Marvin~N Wright.
\newblock {CountARFactuals} -- generating plausible model-agnostic counterfactual explanations with adversarial random forests.
\newblock In {\em Explainable Artificial Intelligence}, pages 85--107, 2024.

\bibitem{redelmeier2024mcce}
Annabelle Redelmeier, Martin Jullum, Kjersti Aas, and Anders L\o{}land.
\newblock {MCCE}: Monte carlo sampling of valid and realistic counterfactual explanations for tabular data.
\newblock {\em Data Mining and Knowledge Discovery}, 38(4):1830–1861, 2024.

\bibitem{zein2022foolXGB}
EL~Hacen Zein and Tanguy Urvoy.
\newblock Tabular data generation: Can we fool {XGB}oost ?
\newblock In {\em NeurIPS 2022 First Table Representation Workshop}, 2022.

\bibitem{bird2024cifake}
Jordan~J Bird and Ahmad Lotfi.
\newblock {CIFAKE}: Image classification and explainable identification of {AI}-generated synthetic images.
\newblock {\em IEEE Access}, 12:15642--15650, 2024.

\bibitem{abir2023deepfakeImageXAI}
Wahidul~Hasan Abir, Faria~Rahman Khanam, Kazi~Nabiul Alam, Myriam Hadjouni, Hela Elmannai, Sami Bourouis, Rajesh Dey, and Mohammad~Monirujjaman Khan.
\newblock Detecting deepfake images using deep learning techniques and explainable {AI} methods.
\newblock {\em Intelligent Automation \& Soft Computing}, 35(2):2151--2169, 2023.

\bibitem{baraheem2023AIvsAI}
Samah~S Baraheem and Tam~V Nguyen.
\newblock {AI} vs. {AI}: Can {AI} detect {AI}-generated images?
\newblock {\em Journal of Imaging}, 9(10):199, 2023.

\bibitem{mohamed2016implicit}
Shakir Mohamed and Balaji Lakshminarayanan.
\newblock Learning in implicit generative models.
\newblock {\em arXiv preprint arXiv:1610.03483}, 2016.

\bibitem{kingma2014auto}
Diederik~P Kingma and Max Welling.
\newblock Auto-encoding variational bayes.
\newblock In {\em International Conference on Learning Representations}, 2014.

\bibitem{goodfellow2014generative}
Ian Goodfellow, Jean Pouget-Abadie, Mehdi Mirza, Bing Xu, David Warde-Farley, Sherjil Ozair, Aaron Courville, and Yoshua Bengio.
\newblock Generative adversarial nets.
\newblock In {\em Advances in Neural Information Processing Systems}, volume~27, 2014.

\bibitem{rezende2015variational}
Danilo Rezende and Shakir Mohamed.
\newblock Variational inference with normalizing flows.
\newblock In {\em Proceedings of the 32nd International Conference on Machine Learning}, volume~37 of {\em PMLR}, pages 1530--1538, 2015.

\bibitem{bengio2000neural}
Yoshua Bengio, R{\'e}jean Ducharme, and Pascal Vincent.
\newblock A neural probabilistic language model.
\newblock In {\em Advances in Neural Information Processing Systems}, volume~13, 2000.

\bibitem{bond2021deep}
Sam Bond-Taylor, Adam Leach, Yang Long, and Chris~G Willcocks.
\newblock Deep generative modelling: A comparative review of vaes, gans, normalizing flows, energy-based and autoregressive models.
\newblock {\em IEEE Transactions on Pattern Analysis and Machine Intelligence}, 44(11):7327--7347, 2022.

\bibitem{xu2019ctgan}
Lei Xu, Maria Skoularidou, Alfredo Cuesta-Infante, and Kalyan Veeramachaneni.
\newblock Modeling tabular data using conditional {GAN}.
\newblock In {\em Advances in Neural Information Processing Systems}, volume~32, 2019.

\bibitem{kotelnikov2023tabddpm}
Akim Kotelnikov, Dmitry Baranchuk, Ivan Rubachev, and Artem Babenko.
\newblock {T}ab{DDPM}: Modelling tabular data with diffusion models.
\newblock In {\em Proceedings of the 40th International Conference on Machine Learning}, volume 202 of {\em PMLR}, pages 17564--17579, 2023.

\bibitem{zhao2023tabula}
Zilong Zhao, Robert Birke, and Lydia Chen.
\newblock Tabula: Harnessing language models for tabular data synthesis.
\newblock {\em arXiv preprint arXiv:2310.12746}, 2023.

\bibitem{nowok2016synthpop}
Beata Nowok, Gillian~M. Raab, and Chris Dibben.
\newblock synthpop: Bespoke creation of synthetic data in {R}.
\newblock {\em Journal of Statistical Software}, 74(11):1–26, 2016.

\bibitem{watson2023adversarial}
David~S Watson, Kristin Blesch, Jan Kapar, and Marvin~N Wright.
\newblock Adversarial random forests for density estimation and generative modeling.
\newblock In {\em Proceedings of The 26th International Conference on Artificial Intelligence and Statistics}, volume 206 of {\em PMLR}, pages 5357--5375, 2023.

\bibitem{qian2023synthcity}
Zhaozhi Qian, Bogdan-Constantin Cebere, and Mihaela van~der Schaar.
\newblock Synthcity: Facilitating innovative use cases of synthetic data in different data modalities.
\newblock {\em arXiv preprint arXiv:2301.07573}, 2023.

\bibitem{fossing2024evaluationSynthcityModels}
Emma F{\"o}ssing and J{\"o}rg Drechsler.
\newblock An evaluation of synthetic data generators implemented in the {P}ython library synthcity.
\newblock In {\em Privacy in Statistical Databases}, pages 178--193, 2024.

\bibitem{grinsztajn2022tree}
L{\'e}o Grinsztajn, Edouard Oyallon, and Ga{\"e}l Varoquaux.
\newblock Why do tree-based models still outperform deep learning on typical tabular data?
\newblock In {\em Advances in Neural Information Processing Systems}, volume~35, pages 507--520, 2022.

\bibitem{borisov2022deep}
Vadim Borisov, Tobias Leemann, Kathrin Seßler, Johannes Haug, Martin Pawelczyk, and Gjergji Kasneci.
\newblock Deep neural networks and tabular data: A survey.
\newblock {\em IEEE Transactions on Neural Networks and Learning Systems}, 35(6):7499--7519, 2024.

\bibitem{shwartz2022tabular}
Ravid Shwartz-Ziv and Amitai Armon.
\newblock Tabular data: Deep learning is not all you need.
\newblock {\em Information Fusion}, 81:84--90, 2022.

\bibitem{hollmann2022tabpfn}
Noah Hollmann, Samuel M{\"u}ller, Katharina Eggensperger, and Frank Hutter.
\newblock Tab{PFN}: A transformer that solves small tabular classification problems in a second.
\newblock In {\em NeurIPS 2022 First Table Representation Workshop}, 2022.

\bibitem{zhangTabSyn}
Hengrui Zhang, Jiani Zhang, Zhengyuan Shen, Balasubramaniam Srinivasan, Xiao Qin, Christos Faloutsos, Huzefa Rangwala, and George Karypis.
\newblock Mixed-type tabular data synthesis with score-based diffusion in latent space.
\newblock In {\em International Conference on Learning Representations}, 2024.

\bibitem{theis2016note}
L~Theis, A~van~den Oord, and M~Bethge.
\newblock A note on the evaluation of generative models.
\newblock In {\em International Conference on Learning Representations}, 2016.

\bibitem{sajjadi2018assessing}
Mehdi~SM Sajjadi, Olivier Bachem, Mario Lucic, Olivier Bousquet, and Sylvain Gelly.
\newblock Assessing generative models via precision and recall.
\newblock In {\em Advances in Neural Information Processing Systems}, volume~31, 2018.

\bibitem{bischoff2024practical}
Sebastian Bischoff, Alana Darcher, Michael Deistler, Richard Gao, Franziska Gerken, Manuel Gloeckler, Lisa Haxel, Jaivardhan Kapoor, Janne~K Lappalainen, Jakob~H Macke, et~al.
\newblock A practical guide to sample-based statistical distances for evaluating generative models in science.
\newblock {\em Transactions on Machine Learning Research}, 2024.

\bibitem{molnar2022interpretable}
Christoph Molnar.
\newblock {\em Interpretable Machine Learning}.
\newblock 2 edition, 2022.

\bibitem{molnar2022pitfalls}
Christoph Molnar, Gunnar K{\"o}nig, Julia Herbinger, Timo Freiesleben, Susanne Dandl, Christian~A. Scholbeck, Giuseppe Casalicchio, Moritz Grosse-Wentrup, and Bernd Bischl.
\newblock General pitfalls of model-agnostic interpretation methods for machine learning models.
\newblock In {\em International Workshop on Extending Explainable AI Beyond Deep Models and Classifiers}, pages 39--68, 2022.

\bibitem{lundberg2017SHAP}
Scott~M Lundberg and Su-In Lee.
\newblock A unified approach to interpreting model predictions.
\newblock In {\em Advances in Neural Information Processing Systems}, volume~30, 2017.

\bibitem{lundberg2018treeSHAP}
Scott~M Lundberg, Gabriel~G Erion, and Su-In Lee.
\newblock Consistent individualized feature attribution for tree ensembles.
\newblock {\em arXiv preprint arXiv:1802.03888}, 2018.

\bibitem{friedman2001PDP}
Jerome~H Friedman.
\newblock Greedy function approximation: a gradient boosting machine.
\newblock {\em Annals of Statistics}, 29(5):1189--1232, 2001.

\bibitem{Goldstein2015ICE}
Justin~Bleich Alex~Goldstein, Adam~Kapelner and Emil Pitkin.
\newblock Peeking inside the black box: Visualizing statistical learning with plots of individual conditional expectation.
\newblock {\em Journal of Computational and Graphical Statistics}, 24(1):44--65, 2015.

\bibitem{Fisher2019PFI}
Aaron Fisher, Cynthia Rudin, and Francesca Dominici.
\newblock All models are wrong, but many are useful: Learning a variable's importance by studying an entire class of prediction models simultaneously.
\newblock {\em Journal of Machine Learning Research}, 20(177):1--81, 2019.

\bibitem{aas2019explaining}
Kjersti Aas, Martin Jullum, and Anders L{\o}land.
\newblock Explaining individual predictions when features are dependent: More accurate approximations to {S}hapley values.
\newblock {\em Artificial Intelligence}, 298:103502, 2021.

\bibitem{chen2023algorithms}
Hugh Chen, Ian~C Covert, Scott~M Lundberg, and Su-In Lee.
\newblock Algorithms to estimate {S}hapley value feature attributions.
\newblock {\em Nature Machine Intelligence}, 5(6):590--601, 2023.

\bibitem{chen2020true}
Hugh Chen, Joseph~D. Janizek, Scott Lundberg, and Su-In Lee.
\newblock True to the model or true to the data?
\newblock {\em arXiv preprint arXiv:2006.16234}, 2020.

\bibitem{simon2019revisiting}
Loic Simon, Ryan Webster, and Julien Rabin.
\newblock Revisiting precision recall definition for generative modeling.
\newblock In {\em Proceedings of the 36th International Conference on Machine Learning}, volume~97 of {\em PMLR}, pages 5799--5808, 2019.

\bibitem{Chen2016XGBoost}
Tianqi Chen and Carlos Guestrin.
\newblock {XGB}oost: A scalable tree boosting system.
\newblock In {\em Proceedings of the 22nd ACM SIGKDD International Conference on Knowledge Discovery and Data Mining}, KDD '16, page 785–794, 2016.

\bibitem{watson2021CPI}
David~S. Watson and Marvin~N. Wright.
\newblock Testing conditional independence in supervised learning algorithms.
\newblock {\em Machine Learning}, 110:2107 -- 2129, 2021.

\bibitem{shapley1953SV}
Lloyd~S Shapley.
\newblock A value for n-person games.
\newblock In Harold~W. Kuhn and Albert~W. Tucker, editors, {\em Contributions to the Theory of Games}, pages 307--317. Princeton University Press, Princeton, 1953.

\bibitem{apley2020ALE}
Daniel~W Apley and Jingyu Zhu.
\newblock Visualizing the effects of predictor variables in black box supervised learning models.
\newblock {\em Journal of the Royal Statistical Society Series B: Statistical Methodology}, 82(4):1059--1086, 2020.

\bibitem{sundararajan2020taylorInteract}
Mukund Sundararajan, Kedar Dhamdhere, and Ashish Agarwal.
\newblock The shapley taylor interaction index.
\newblock In {\em Proceedings of the 37th International Conference on Machine Learning}, volume 119 of {\em PMLR}, pages 9259--9268, 13--18 Jul 2020.

\bibitem{guidotti2024counterfactual}
Riccardo Guidotti.
\newblock Counterfactual explanations and how to find them: literature review and benchmarking.
\newblock {\em Data Mining and Knowledge Discovery}, 38(5):2770--2824, 2024.

\bibitem{adult_data}
Barry Becker and Ronny Kohavi.
\newblock {Adult}.
\newblock UCI Machine Learning Repository, 1996.

\bibitem{nursery_data}
Vladislav Rajkovic.
\newblock {Nursery}.
\newblock UCI Machine Learning Repository, 1989.

\bibitem{zhao2024ctab}
Zilong Zhao, Aditya Kunar, Robert Birke, Hiek Van~der Scheer, and Lydia~Y Chen.
\newblock {CTAB-GAN+}: Enhancing tabular data synthesis.
\newblock {\em Frontiers in Big Data}, 6:1296508, 2024.

\bibitem{UCI_Repo}
Dheeru Dua and Casey Graff.
\newblock Uci machine learning repository, 2017.

\bibitem{kaggle_repo}
Kaggle.
\newblock Kaggle datasets repository.
\newblock Accessed: 2025-02-19.

\bibitem{breiman2001RF}
Leo Breiman.
\newblock Random forests.
\newblock {\em Machine learning}, 45:5--32, 2001.

\bibitem{fumagalli2025unifying}
Fabian Fumagalli, Maximilian Muschalik, Eyke H{\"u}llermeier, Barbara Hammer, and Julia Herbinger.
\newblock Unifying feature-based explanations with functional {ANOVA} and cooperative game theory.
\newblock In {\em The 28th International Conference on Artificial Intelligence and Statistics}, 2025.

\bibitem{redelmeier2020}
Annabelle Redelmeier, Martin Jullum, and Kjersti Aas.
\newblock Explaining predictive models with mixed features using shapley values and conditional inference trees.
\newblock In {\em ICross-Domain Conference for Machine Learning and Knowledge Extraction}, pages 117--137, 2020.

\end{thebibliography}

\newpage

\appendix
\label{ap}



\section{Answering \ref{enum:Q1} -- \ref{enum:Q4} for Synthetic Nursery Data}
\label{ap:nursery}

\begin{figure}[h!]
    \centering
    \includegraphics[width=\linewidth]{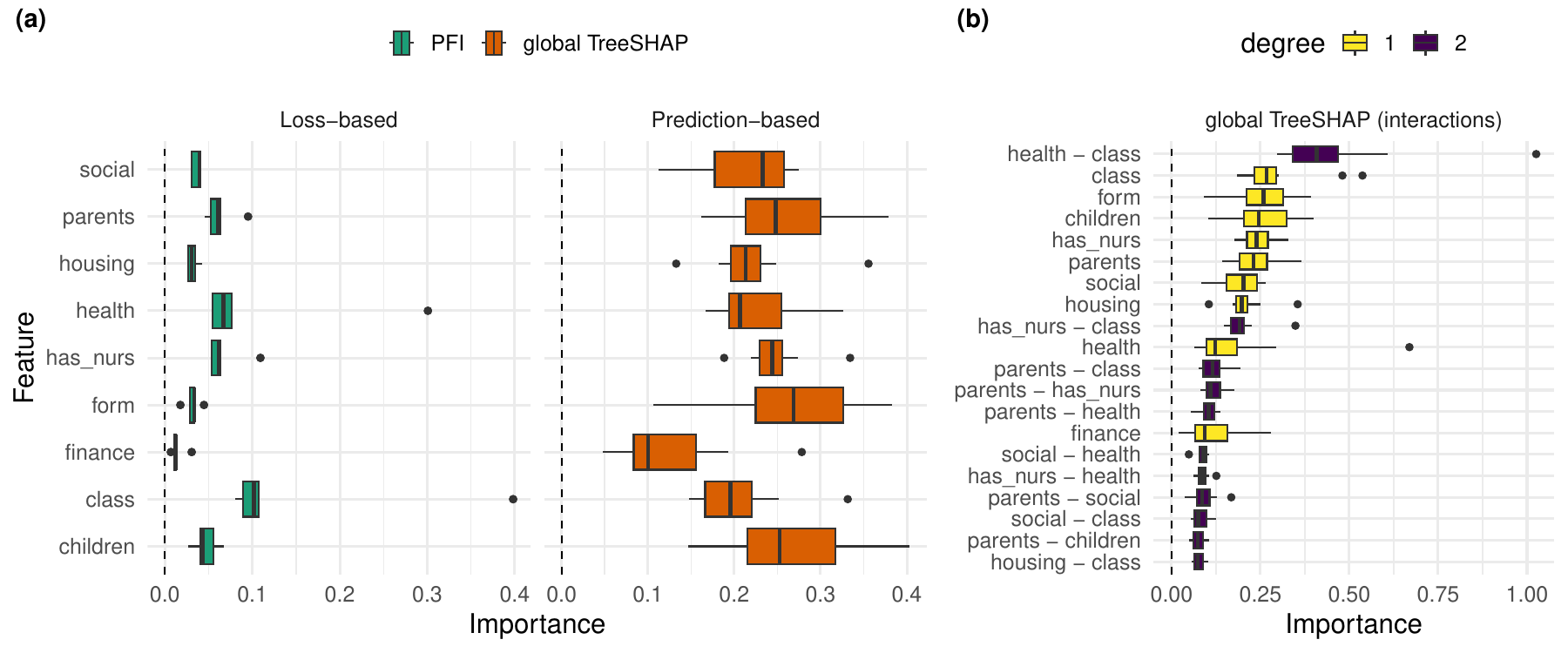}
    \caption{Feature importance values for synthetic data detection with XGBoost for ten CTGAN-generated synthetic nursery datasets. Higher importance values indicate poorer synthesis quality. (a) PFI and global TreeSHAP values. (b) Global TreeSHAP interaction values of degree 1 and 2 (top 20 most important)}
    \label{fig:q1_nursery}
\end{figure}
\begin{figure}[h!]
    \centering
    \includegraphics[width=\linewidth]{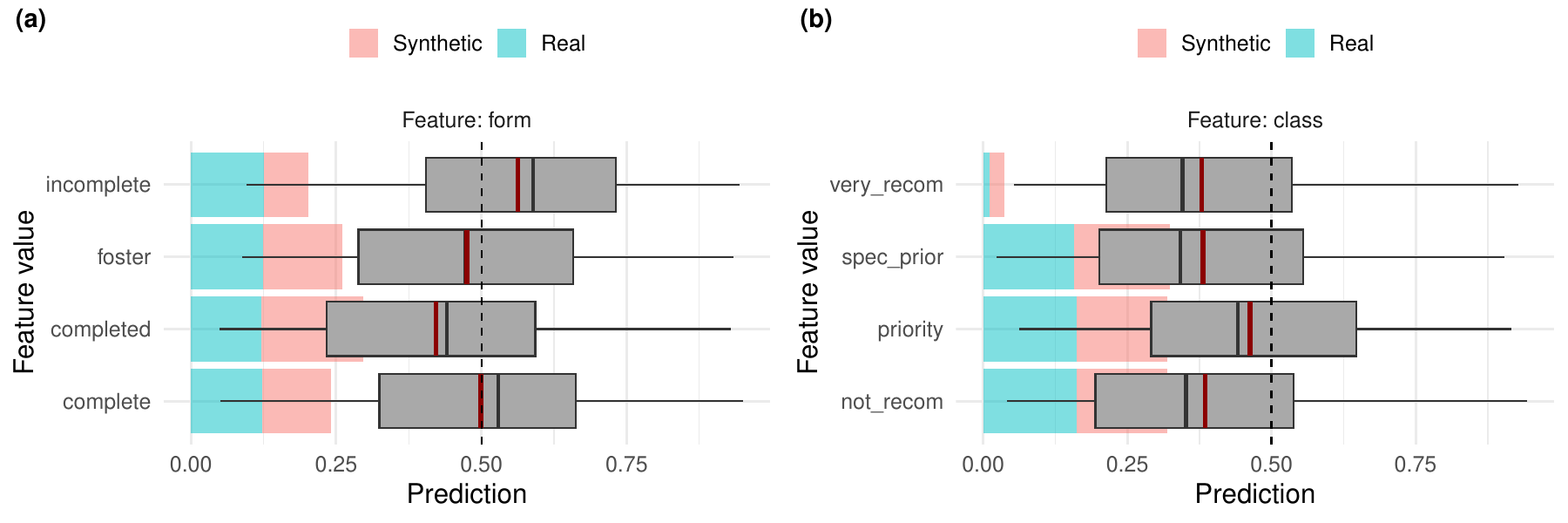}
    \caption{ICE/PDP for synthetic data detection with XGBoost for CTGAN-generated synthetic nursery data. (a) Feature \texttt{form}, (b) feature \texttt{class}. PDP in red, frequencies for real and synthetic data on y-axis.}
    \label{fig:q2_nursery}
\end{figure}

We provide another exemplary application of our approach using the nursery dataset and the generative model CTGAN \cite{xu2019ctgan}. The nursery dataset has 12\,958 instances and 9 columns (all categorical). CTGAN is a widely used GAN-based generative model for tabular data. Again, on each of these generated datasets, we trained an XGBoost model for synthetic data detection using Bayesian optimization for hyperparameter tuning and a train-test split of 30\%, while keeping the real data points the same across all ten synthetic datasets. The mean accuracy and its variation on both train and test data are shown in Figure~\ref{fig:detection_performance}b: The accuracy is consistently around 70\%, with slightly lower values on test data.

Figure~\ref{fig:q1_nursery} shows that CTGAN struggled to accurately reproduce multiple features. On top of that, especially the dependency between \texttt{health} and \texttt{class} appears to be insufficiently retained. In alignment with Figure~\ref{fig:q1_nursery}b, the ICE and PDP values for the feature \texttt{form} in Figure~\ref{fig:q2_nursery}a indicate a poor reconstruction of the marginal distribution with over- and underrepresented classes. The situation differs for the feature \texttt{class}, as shown in Figure~\ref{fig:q2_nursery}: Except for the rare class \texttt{very\_recom}, the classes appear to be equally represented in real and synthetic data. The low ICE and PDP values for the three frequent classes are therefore likely due to inadequately captured dependencies with other features,
consistent with the presence of multiple relevant interactions for \texttt{class} in Figure~\ref{fig:q1_nursery}b.

\begin{figure}[t]
    \centering
    \includegraphics[width=\linewidth]{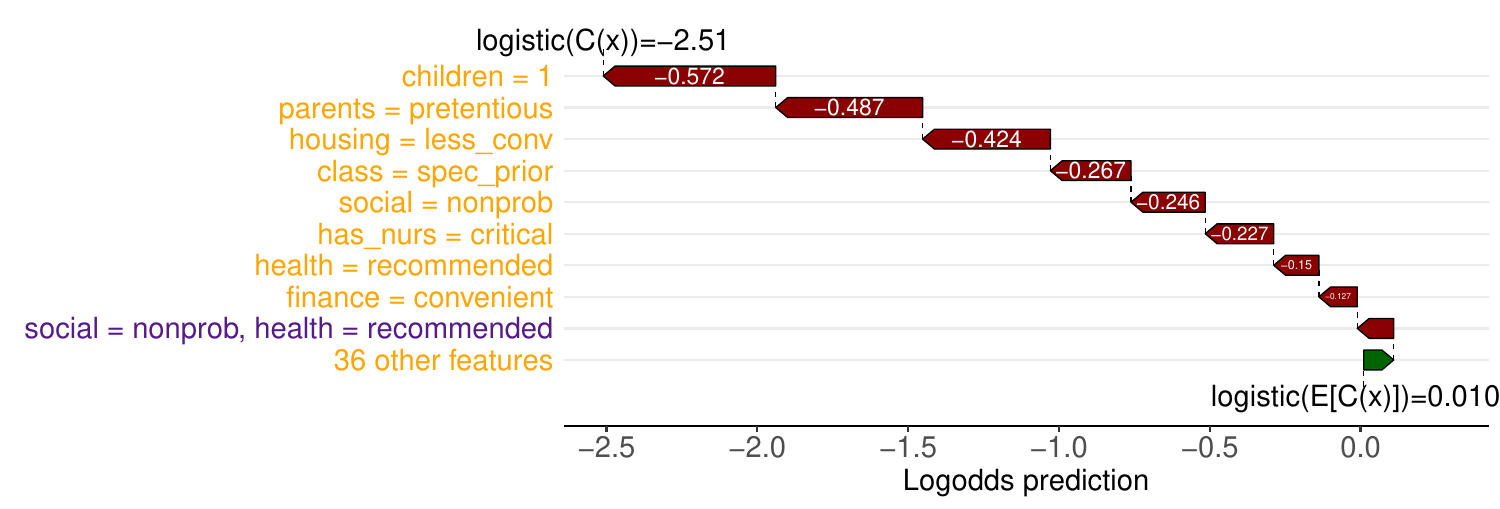}
    \caption{Waterfall plots for Shapley interaction values decomposing the XGBoost prediction for an examplary instance of CTGAN-generated synthetic nursery data. Note that TreeSHAP decomposes the prediction on the log-odds/logistic scale rather than on the probability scale.}
    \label{fig:q3_nursery}
\end{figure}

\begin{figure}[t]
    \centering
    \includegraphics[width=\linewidth]{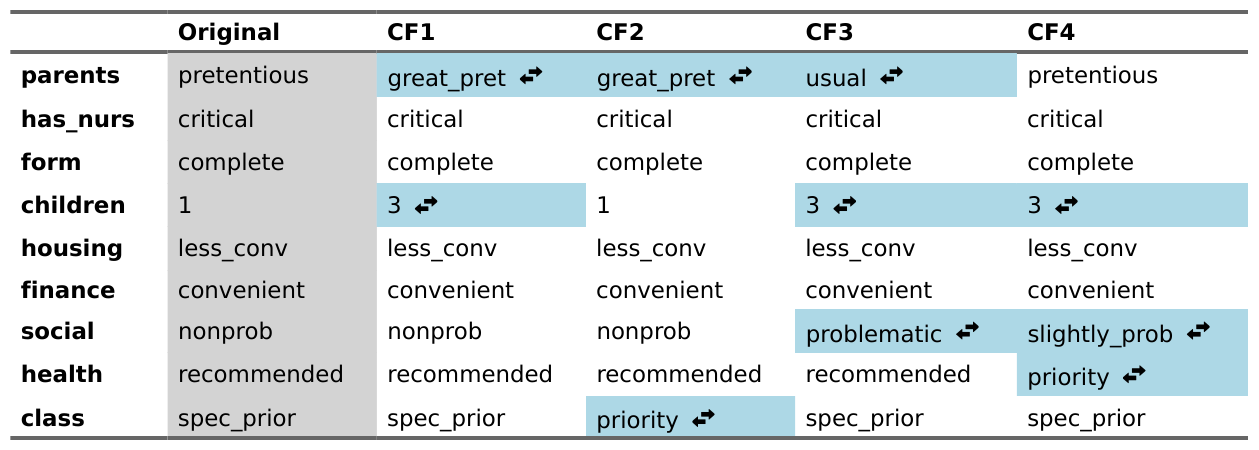}
    \caption{Four counterfactual explanations for an examplary instance of CTGAN-generated synthetic nursery data. Highlighted features are changed.}
    \label{fig:q4_nursery}
\end{figure}

Finally, we consider a specific correctly classified ($C(\mathbf{x}) = 0.07$) synthetic observation with pretentious parents, one child and non-problematic social conditions, to mention some key features.
The Shapley interaction values of this prediction, displayed in Figure~\ref{fig:q3_nursery}, indicate that correct classification is mainly driven by single features as opposed to feature interactions. The feature values \texttt{children} $= 1$ and \texttt{parents} = \texttt{pretentious} contribute the most, but nearly all other features also have a substantial impact, implying that the majority of the features values are unrealistic or overrepresented.
Figure~\ref{fig:q4_nursery} shows four counterfactual examples for the same individual. Modifying the \texttt{parents} feature and changing the \texttt{children} feature to 3 are commonalities in three of the examples. 
However, like in Section~\ref{sec:ex_q1-4}, the synthetic observation can be altered in quite different ways to appear realistic to the detection model.

\end{document}